Université de Montréal

**Evolving Artificial Neural Networks
To Imitate Human Behaviour
In Shinobi III : Return of the Ninja Master**

par

**Maximilien Le Cleï**

Département d'informatique et de recherche opérationnelle, Faculté des arts et des sciences

Mémoire présenté en vue de l'obtention du grade de Maîtrise ès sciences (M.Sc.)
en Informatique, option Intelligence Artificielle

Août 2021

# Université de Montréal

Département d'informatique et de recherche opérationnelle, Faculté des arts et des sciences

*Ce mémoire intitulé*

**Evolving Artificial Neural Networks
To Imitate Human Behaviour
In Shinobi III : Return of the Ninja Master**

*Présenté par*

**Maximilien Le Cleï**

*A été évalué par un jury composé des personnes suivantes*

**Ioannis Mitliagkas**
Président-rapporteur

**Pierre-Louis Bellec**
Directeur de recherche

**Devon Hjelm**
Codirecteur

**Sarath Chandar Anbil Parthipan**
Membre du jury

# Résumé

Notre société est de plus en plus friande d'outils informatiques. Ce phénomène s'est particulièrement accru lors de cette dernière décennie suite, entre autres, à l'émergence d'un nouveau paradigme d'Intelligence Artificielle. Plus précisément, le couplage de deux techniques algorithmiques, les Réseaux de Neurones Profonds et la Descente de Gradient Stochastique, propulsé par une force de calcul exponentiellement croissante, est devenu et continue de devenir un atout majeur dans de nombreuses nouvelles technologies. Néanmoins, alors que le progrès suit son cours, certains se demandent toujours si d'autres méthodes pourraient similairement, voire davantage, bénéficier de ces diverses avancées matérielles.

Afin de pousser cette étude, nous nous attelons dans ce mémoire aux Algorithmes Évolutionnaires et leur application aux Réseaux de Neurones Dynamiques, deux techniques dotées d'un grand nombre de propriétés avantageuses n'ayant toutefois pas trouvé leur place au sein de l'Intelligence Artificielle contemporaine. Nous trouvons qu'en élaborant de nouvelles méthodes tout en exploitant une forte puissance informatique, il nous devient alors possible de développer des agents à haute performance sur des bases de référence ainsi que d'autres agents se comportant de façon très similaire à des sujets humains sur le jeu vidéo *Shinobi III : Return of The Ninja Master*, cas typique de tâches complexes que seule l'optimisation par gradient était capable d'aborder jusqu'alors.

**Mots clés**

Intelligence Artificielle, Algorithmes Évolutionnaires, Réseaux de Neurones Artificiels

# Abstract

Our society is increasingly fond of computational tools. This phenomenon has greatly increased over the past decade following, among other factors, the emergence of a new Artificial Intelligence paradigm. Specifically, the coupling of two algorithmic techniques, Deep Neural Networks and Stochastic Gradient Descent, thrusted by an exponentially increasing computing capacity, has and is continuing to become a major asset in many modern technologies. However, as progress takes its course, some still wonder whether other methods could similarly or even more greatly benefit from these various hardware advances.

In order to further this study, we delve in this thesis into Evolutionary Algorithms and their application to Dynamic Neural Networks, two techniques which despite enjoying many advantageous properties have yet to find their niche in contemporary Artificial Intelligence. We find that by elaborating new methods while exploiting strong computational resources, it becomes possible to develop strongly performing agents on a variety of benchmarks but also some other agents behaving very similarly to human subjects on the video game *Shinobi III : Return of The Ninja Master*, typical complex tasks previously out of reach for non-gradient-based optimization.

# Contents

# List of Tables

# List of Figures

# List of Acronyms & Abbreviations

**DCN**  Dynamic Convolutional Network

**DRN**  Dynamic Recurrent Network

**GA**  Genetic Algorithm

**NEAT**  NeuroEvolution of Augmenting Topologies

**RL**  Reinforcement Learning

# Context

## 1. Artificial Neural Networks

Artificial Neural Networks (McCulloch & Pitts, 1943; Farley & Clark, 1954) are computing techniques inspired by biological nervous systems. Just like their biological counterpart, these artificial systems are composed of interconnected neurons that receive, process and transmit information signals (Figure 1).

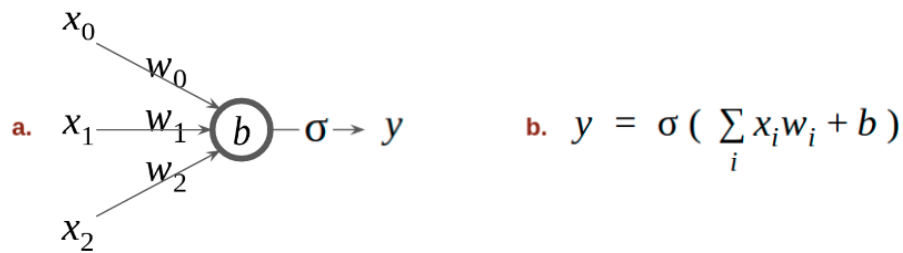

*Figure 1.* **Artificial Neuron.** (a) This neuron inputs values ($x_0$, $x_1$, $x_2$), which it respectively multiplies with weights ($w_0$, $w_1$, $w_2$), then sums the products and bias **b** down to a single value, now transformed by non-linear function **σ(x)** to finally output **y**. For instance, with respective values (**1**, **2**, **3**), (**-1**, **1**, **0**), **-0.5** and **σ(x) = max(x, 0)**, this neuron outputs **y = 0.5**. (b) Generalized mathematical formula.

### 1.1. Feedforward Neural Networks

Feedforward Neural Networks (Rosenblatt, 1958; Widrow & Hoff, 1960) are some of the most common types of artificial neural networks. These networks make use of a layered directed structure through which they process scalar information acyclically. The first layer is made up of so-called **input nodes** that receive information signals from an outside source only to propagate them, without transformation, to the network's next layer of nodes. The last layer is made up of **output nodes** that receive, transform and transfer information back out of the network. The simplest instance of such networks is the two-layer feedforward neural network wherein output nodes transform information directly received from input nodes (Figure 2).

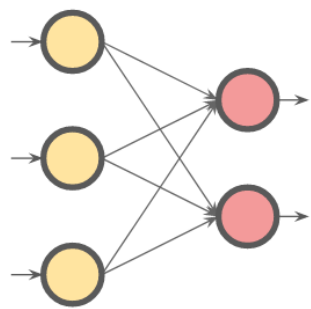

*Figure 2.* **Two-Layer Feedforward Neural Network.** Input nodes each propagate one of the values fed into the system. Output nodes receive and transform information provided by each of the input nodes and feed it out of the system.

In order to model more complex functions, additional layers of **hidden nodes** ought to be positioned between input and output nodes (Figure 3). These nodes can be utilized as transitory computing units in order to send more sophisticated information signals to nodes further down the network.

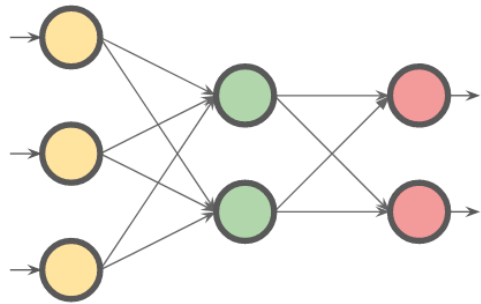

*Figure 3*. **Three-Layer Feedforward Neural Network.** In between input and output nodes can be inserted hidden nodes in order to complexify a network's computational graph and therefore allow it to model more complex functions.

### 1.2. Recurrent Neural Networks

Due to the fully forward-flowing nature of their computational graphs, Feedforward Neural Networks do not retain any of the information they compute. Recurrent Neural Networks (Hopfield, 1982; Rumelhart, Hinton & Williams, 1985) are extensions to these networks wherein certain nodes, or layers of nodes, connect backward in the computational graph, usually onto themselves, allowing some of the previously computed information to be fed back into the system for future network passes (Figure 4).

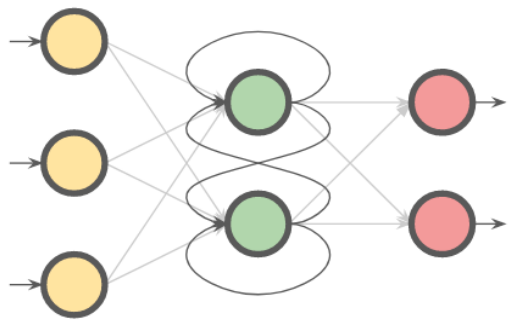

*Figure 4*. **Recurrent Neural Network.** A simple recurrent neural network wherein hidden nodes connect back onto themselves, meaning that information computed by hidden nodes at time **t** is fed back to these very nodes at time **t+1**.

### 1.3. Convolutional Neural Networks

The different types of Artificial Neural Networks discussed thus far input and process sets of scalar values. Image data however, most often structured as a grid of RGB[1] pixel values, cannot properly be transformed into such disposition without sacrificing its structural information.

---

[1] Red Green Blue

Inspired by work on animal vision, 福島邦彦, 1979 first proposed the "Neocognitron", an Artificial Neural Network making use of two new types of operations called the convolution (Figure 5) and pooling (Figure 6) operations. This class of networks, now most often referred to as Convolutional Neural Networks (Figure 7), have become, in combination with the backpropagation algorithm (LeCun et al., 1989), the standard neural network technique to process image data.

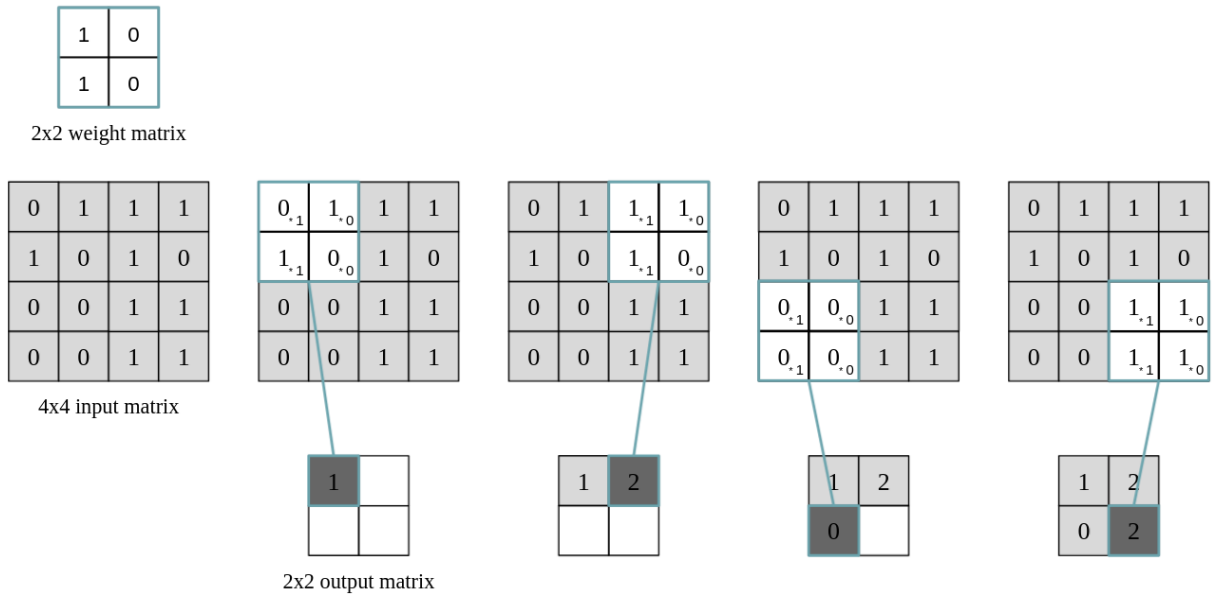

*Figure 5.* **Convolution Operation Example.** A 2x2 weight kernel is convolved over an input image, multiplying then summing down values, producing a corresponding output matrix.

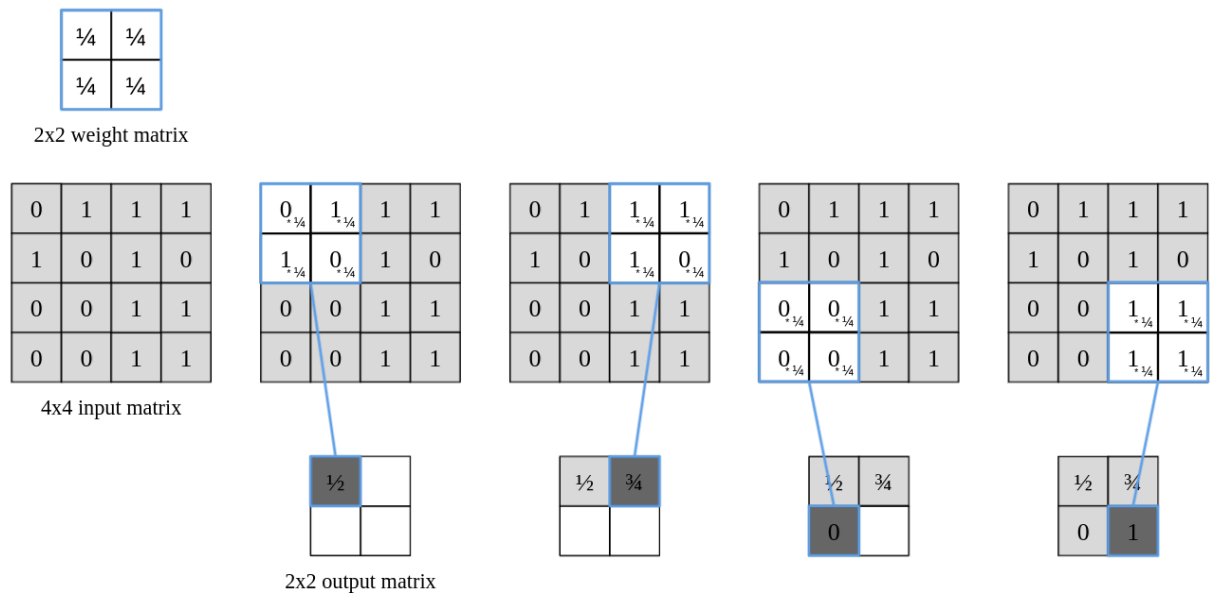

*Figure 6.* **Pooling Operation Example.** The (average) pooling operation can be framed as a special case of the convolution operation wherein kernel weights are equal and sum to 1.

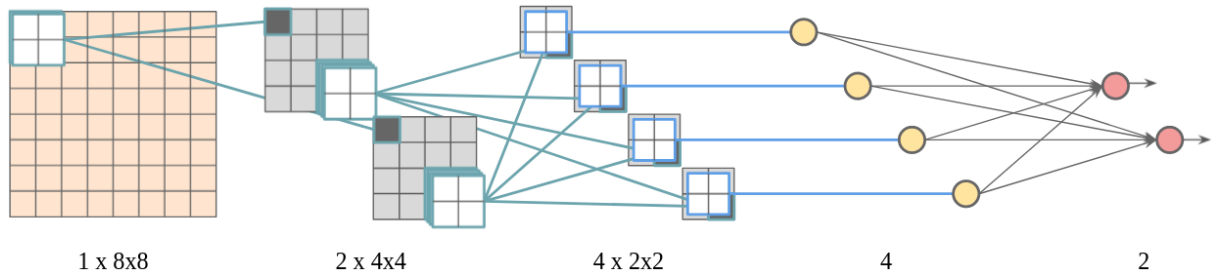

1 x 8x8     2 x 4x4     4 x 2x2     4     2

*Figure 7.* **Convolutional Neural Network Example.** This neural network takes as input a single-channel 8x8 image. Two weight filters are convolved over it producing two 4x4 feature maps. These two feature maps are each convolved over by four new weight kernels, producing four 2x2 feature maps. A 2x2 average pooling operation is then applied, producing four scalar values fed into four traditional neurons. Finally, these four scalar values are transformed through two output nodes using standard artificial neuron operations described in Figure 1.

## 2. Evolutionary Algorithms

Evolutionary algorithms (Barricelli, 1954; Fraser, 1957; Fogel, Owens & Walsh, 1966; Rechemberg, 1973; Schwefel, 1974; Holland, 1975) are computational techniques inspired by the biological process of natural selection. These algorithms exist under many different forms; however, generically, a population $P$ composed of $N$ agents cycles through three stages: **variation**, **evaluation** and **selection**. During the **variation** stage, agents in the population are altered through two types of operations: **crossovers**, wherein existing agents are combined in order to form new ones and **mutations**, wherein existing agents are randomly perturbed. Next, during the **evaluation** stage, agents are assigned a **fitness** score according to their performance on some predetermined task. Finally, during the **selection** stage, a set of agents are chosen according to their fitness score in order to make up the next generation of the population.

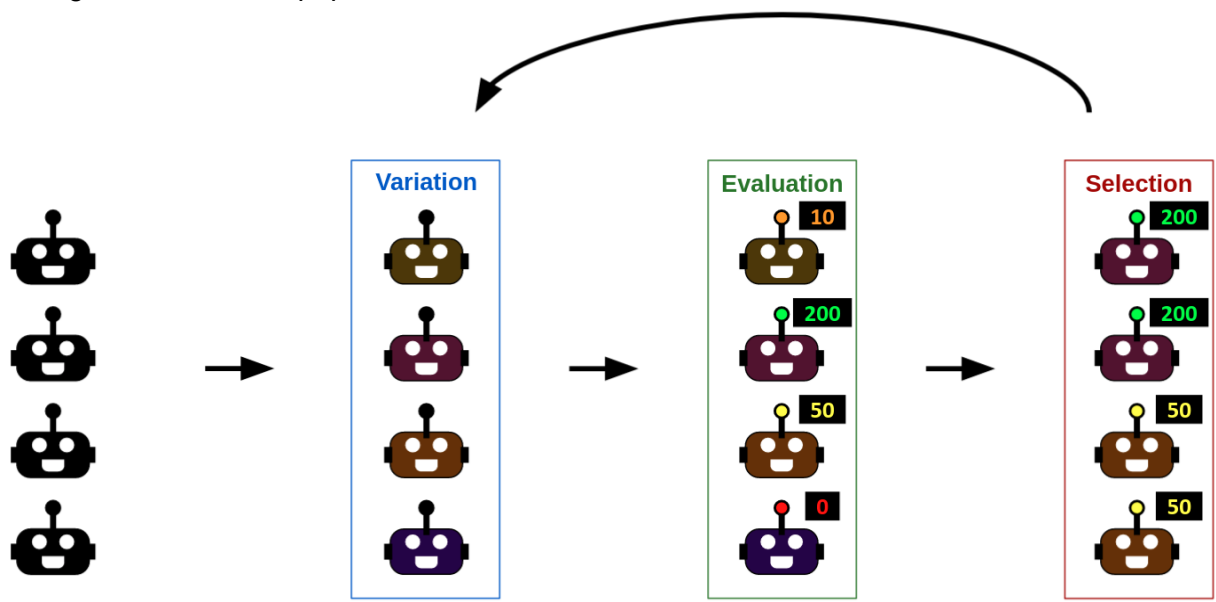

*Figure 8.* **Evolutionary Algorithm Example.** A population is initialized with four identical agents (most left). These four agents are first randomly mutated during the variation stage, receive a fitness score on a given task as a result of the evaluation stage and finally go through a selection stage wherein the top scoring half of the population is maintained and duplicated over the lower half. This process ought to loop back onto itself, repeating indefinitely unless a termination criterion is put in place.

## 2.1. Application to Artificial Neural Networks

By virtue of their loose specification, evolutionary algorithms admit a broad range of use-cases. Amongst popular ones is their application to Artificial Neural Networks (Montana & Davis, 1989; Miller, Todd & Hedge, 1989; Fogel, Fogel & Porto, 1990; Belew, McInerney & Schraudolph, 1990; Whitley, Starkweather & Bogart, 1990), often shortened to Neuroevolution. Neural networks composing artificial agents, during the variation stage, ought to be similarly be perturbed and recombined. These operations can be made to affect anything from connection strengths to network architecture or even meta-level parameters.

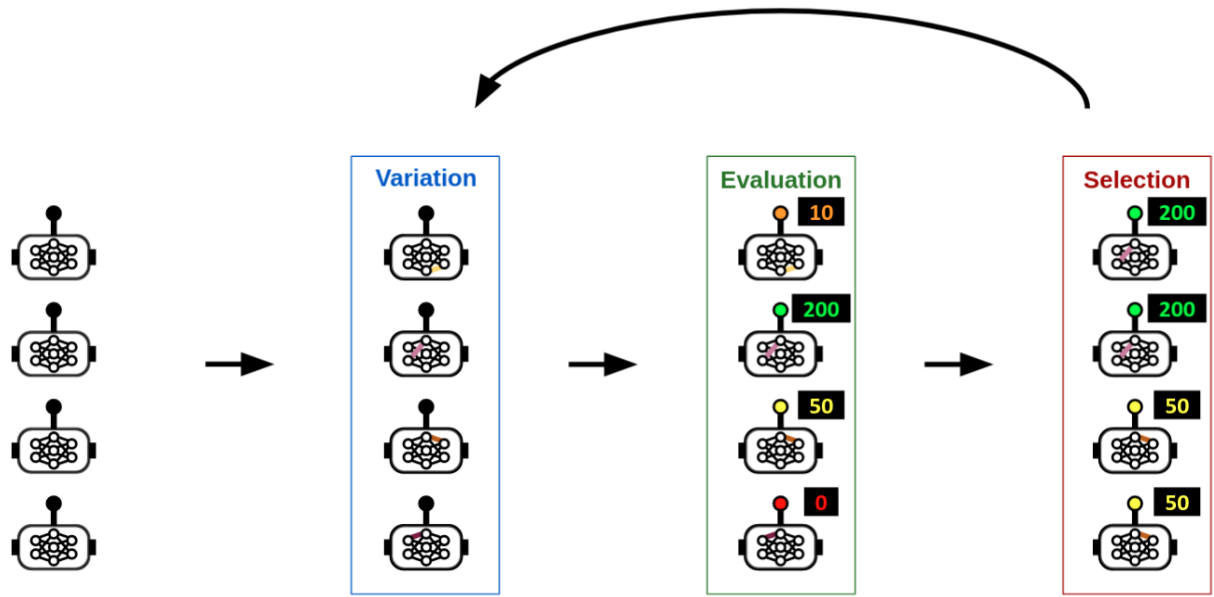

*Figure 9.* **Neuroevolution Example.** We follow the same principles presented in Figure 8 but add in the fact that artificial agents are now equipped with artificial neural networks. In this example, agents are first initialized with identical feedforward neural networks which, during variation stages, see one of their connections randomly perturbed. All these slightly different agents are then evaluated and selected as showcased previously.

## 3. Optimization

### 3.1. Metric Optimization

Historically, the most common way to obtain artificial "learning" systems[2] that produce useful behaviour has been to set-up a quantitative performance criterion, acting as a proxy for an intended qualitative behaviour, and then make use of an algorithm to optimize a system on that metric. For instance, pattern recognition systems are usually evaluated on, and optimized for classification rate, coefficient of determination or even perplexity while control and game playing systems are usually targeted on score and/or completion.

However, many desired qualitative behaviours, such as imitation, have been found quite hard to properly proxy quantitatively.

---

[2] Systems that are set up to be automatically and incrementally optimized (rather than pre-programmed) for a task at hand.

## 3.2. Generative Adversarial Optimization

One tentative workaround to quantitatively proxying for imitation is proposed by the framework of generative adversarial optimization, recently made popular in the context of gradient optimization of neural networks (Goodfellow et al., 2014). In essence, instead of manually quantifying performance, this framework introduces a second agent optimized to discriminate between real collectable data and artificially produced data. Oppositely, the original data-producing agent is now optimized to generate data that the new agent cannot easily discern. When optimization conditions are right, generating agents eventually produce synthesized data so similar to real data that discriminator agents can no longer tell the difference.

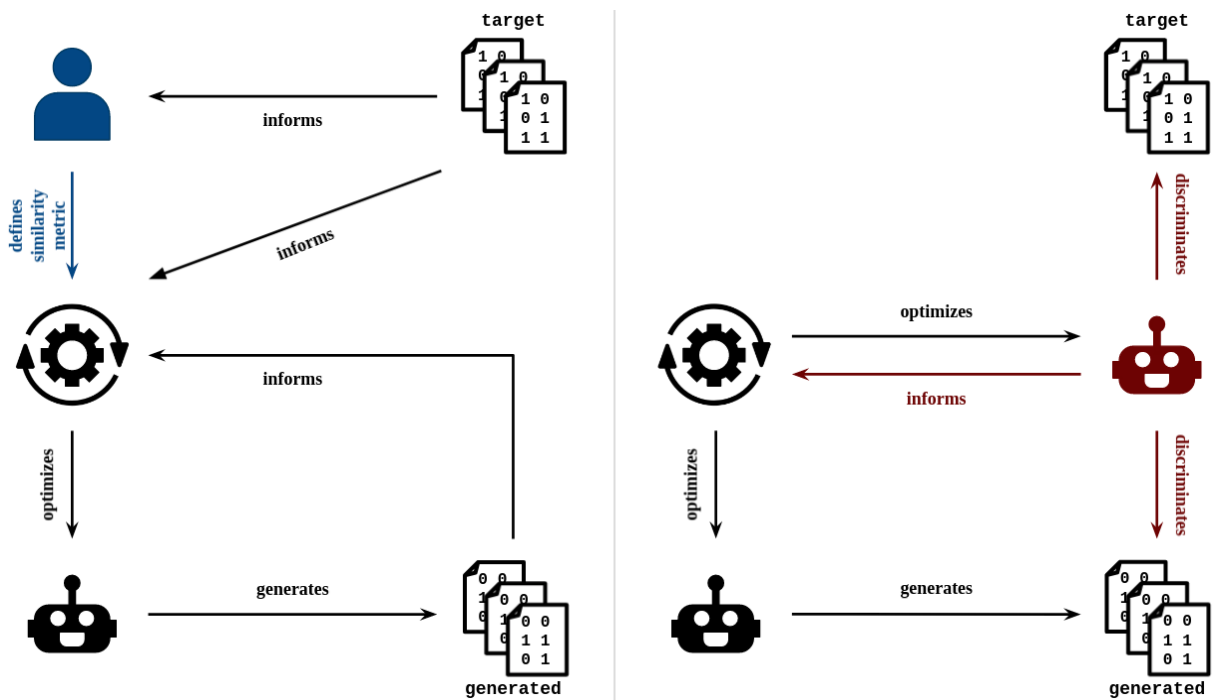

*Figure 10.* **Metric-Based and Adversarial Optimization for Data Imitation.** (left) A similarity metric is manually defined from some target dataset for an agent to be optimized based on. (right) Instead of a metric, a second agent is put in place and optimized to discriminate between generated and target data. The generating agent is now optimized to lower successful discrimination from the new agent.

# Evolving Artificial Neural Networks to Imitate Human Behaviour in Shinobi III: Return of the Ninja Master

Maximilien Le Cleï [1,2], François Paugam [1,2], Olexa Bilaniuk [2], Julie Boyle [1], Yann Harel [1], Paul-Henri Mignot [1], Basile Pinsard [1], Loïc Tetrel [1], Devon Hjelm [2], Pierre Bellec [1]

[1] CRIUGM, Montréal
[2] Mila, Université de Montréal

We attempt to further demonstrate the ability for evolutionary algorithms to leverage modern computational resources. By a) utilizing an adversarial generation framework, b) developing artificial neural networks of dynamic complexity, and c) finding ways to better exploit modern computing systems, we find ourselves able to produce artificial agents with quasi-identical behaviour from human subjects on short temporal windows, in a video game setting.

## 1. Introduction

Gradient-based optimization of Deep Neural Networks (LeCun, Bengio & Hinton, 2015; Schmidhuber, 2015), by virtue of its ability to leverage modern computational resources, has established itself as this decade's single most influential line of work in Artificial Intelligence. However, given the exponentially falling cost of computation, it seems rather likely that other techniques with similarly strong scaling properties will eventually become of interest. An increasingly strong case is being made for evolutionary computation to be one of these. Indeed, when provided sufficient computation, simple evolutionary algorithms paired with deep networks have begun to rival popular modern methods (Salimans et al., 2017; Such et al., 2017; Risi & Stanley, 2019).

In this work, we attempt to further this effort through three contributions. First, we propose an evolutionary algorithm-focused adaptation of the adversarial generation framework recently made popular by Goodfellow et al., 2014. Next, we revisit the dynamic complexification of neural networks, component key to the field's prior success (Stanley & Miikkulainen, 2002). Finally, we present a highly scalable inter-process communication protocol, designed for evolutionary algorithms to better utilize modern computing systems.

We find that these techniques, once provided adequate resources, are able on short time-frames to produce artificial behaviour almost indistinguishable from human subjects in a video game environment.

## 2. Background

Video games have long been used by Artificial Intelligence researchers in order to evaluate their systems' capacities. To this day, most notable work in this area has been produced through the psychology-inspired field of Reinforcement Learning (Sutton & Barto, 2018). This discipline first found success on simple relatively low-information games before successfully merging with Deep Neural Networks (Mnih et al., 2013), enabling it to tackle much higher-dimensional games requiring more complex information processing. Since then, Deep Reinforcement Learning has produced some of the most impressive modern Artificial Intelligence work to date, with agents capable of beating some of the world's best Go (Silver et al., 2016), Dota 2 (Berner et al., 2019) and Starcraft 2 (Vinyals et al., 2019) players.

The story of Neuroevolution (Stanley et al., 2019), name coined for the application of Evolutionary Algorithms to Artificial Neural Networks, is also deeply rooted in video games. To this day, it is perhaps best known for NEAT (Stanley & Miikkulainen, 2002), an algorithm successful in combining various genetic operators in order to build increasingly complex neural networks capable of taking on relatively low-dimensional games. Similarly, the field was recently rekindled following the discovery that very simple evolutionary algorithms could also, given sufficient computational resources, optimize Deep Neural Networks to play complex games (Salimans et al., 2017; Such et al., 2017; Risi & Stanley, 2019). And while the field has yet to produce world champion beating agents, the case has been made that evolutionary components have played a key role in one of these state-of-the-art systems (Arulkumaran, Cully & Togelius, 2019).

Imitation Learning (Hussein et al., 2017) is yet another component that has been essential in creating world champion beating systems. Techniques from this class of work were extensively used to facilitate and speed up many of these artificial agents' skill acquisition by learning from human behaviour. This framework currently exists under many different forms, one of them resulting from a pairing (Ho & Ermon, 2016) with Generative Adversarial Networks (Goodfellow et al., 2014). Finally, while Neuroevolution has been extensively explored both in relation with adversarial frameworks (Stanley & Miikkulainen, 2004; Costa et al., 2020) and human behaviour imitation (Ki, Lyu & Oh, 2006; Miranda, Sanchez-Ruiz & Peinado, 2016; Jalali et al., 2019), the combination of the two appears to this day, relatively unexplored.

## 3. Methods

### 3.1. Evolutionary Adversarial Generation

Evolutionary Algorithms are classical population-based optimization algorithms inspired by the biological process of natural selection. These techniques generally revolve around the idea of setting up a population of agents through an iterative three-stage cycle. Agents first undergo "variation", stage during which they are altered through the use of genetic operators. This stage is followed by an "evaluation" stage, wherein agents are assessed on their ability to perform the task at hand. Finally comes "selection", at which point it is decided which agents will form the basis of the population's next generation.

There are many ways to go about defining each of these stages, however, evaluating agents predominantly relies on defining a quantitative metric proxying for a qualitative objective, called a fitness function. Many tasks however, are fundamentally hard to properly quantify (and others that seemingly aren't are prone to unexpected behaviour, Amodei & Clark, 2016). To circumvent this issue, we propose to set up an adversarial generation framework wherein two equal sized populations of agents match against each other. One population consists of agents generating data points, called the generators, while a second population contains agents, called the discriminators, tasked to dissociate between generated data points and data points sampled from a target dataset. Every iteration, in preparation for the evaluation stage, every discriminator is randomly assigned to one distinct generator and one random "real" data point (Figure 11). In each of these matches, once generators have generated a data point, discriminators are evaluated based on their ability to correctly identify both generated and real data while in turn, generators are evaluated based on their paired discriminator's misidentification of the generated data (Figure 12). Finally, the top 50% agents in each population are selected and duplicated over the respective lower halves.

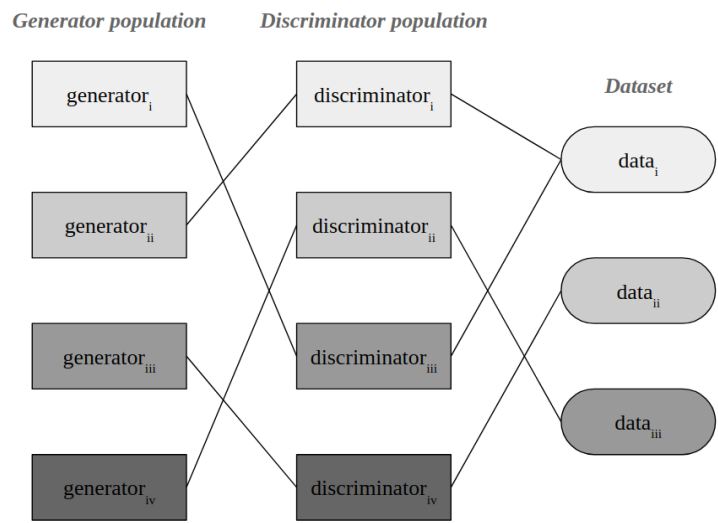

*Figure 11.* **Evolutionary Adversarial Generation : Agent Matching.** Before proceeding with evaluation, each discriminator is randomly assigned a unique generator and a random data point.

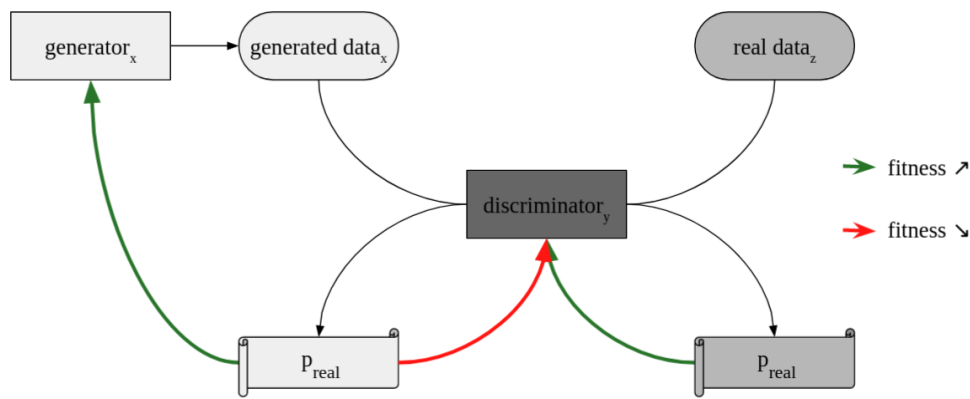

*Figure 12.* **Evolutionary Adversarial Generation : Agent Evaluation.** Once the generator has produced its data point, the discriminator observes both real and generated data points, outputting a score for both, representing its confidence in their authenticity. The generator's fitness is made equal to the discriminator's (incorrect) confidence in the generated data while the discriminator's fitness corresponds to its (incorrect) confidence in the generated data subtracted from its (correct) confidence in the real data. It is to note that this framework does not impose restrictions on data type.

### 3.2. Dynamic Neural Networks

Artificial Neural Networks are biologically inspired computing systems that have become central to Artificial Intelligence's recent accomplishments. Through waves of innovation, these networks have gotten bigger, more efficient, and increasingly competent on tasks like image classification (Tan & Le, 2019), image generation (Karras, Laine & Aila, 2019) and natural language processing (Brown et al., 2020). However, due to the nature of stochastic gradient descent, algorithm through which they are predominantly optimized, this increase in complexity has had to be produced manually. In contrast, Neuroevolution, through its ability to modify both weights and architecture, as successfully showcased in NEAT (Stanley & Miikkulainen, 2002), gives way to a more natural path towards increases in complexity. However, the past decade has been marked by the emergence of higher-dimensional data which neural networks of dynamic complexity have yet to succeed as well on.

In this section, we propose to revisit the topic of dynamic networks by leveraging convolution operations, key components found in most modern Artificial Intelligence systems capable of handling image and video data. We first introduce a Dynamic Recurrent Network (DRN) before following up with a Dynamic Convolutional Network (DCN) and finish by explaining how they combine in a third subsection.

### 3.2.1. Dynamic Recurrent Network

We first present a Dynamic Recurrent Network (DRN), a network structured as a directed layered graph wherein nodes (Figure 13) process information using linear transformations followed by non-linear activation functions. This network evolves in complexity according to four structural mutations:

1) Grow Connection (Section 3.2.1.1.)
2) Prune Connection (Section 3.2.1.2.)
3) Grow Node (Section 3.2.1.3.)
4) Prune Node (Section 3.2.1.4.)

In order to describe these four mutations, we define the following:
**In-nodes:** *(≠ input nodes)* Set of nodes that a node receives information from.
**Out-nodes**: *(≠ output nodes)* Set of nodes that a node emits information to.
**Receiving nodes**: Set comprised of all input nodes + nodes possessing in-nodes.
**Emitting nodes**: List of nodes possessing out-nodes. Nodes appear once per out-node.

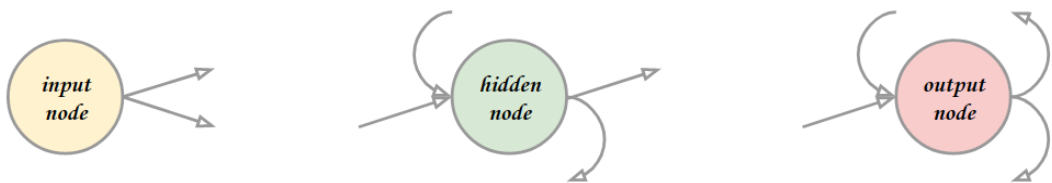

*Figure 13*. **Dynamic Recurrent Network Nodes.** In the first layer, input nodes transmit input information to hidden and output nodes. In the last layer, output nodes receive/emit information from/to potentially any node. In between, hidden nodes, just like output nodes, receive/emit information from/to potentially any node. In the case of forward connections between nodes, nodes on the receiving end obtain and process the emitting node's output information during the same network pass, whereas, in the case of backward/same-layer connections, nodes on the receiving end obtain the emitting nodes's output information during the next pass.

### 3.2.1.1. Mutation #1 : Grow Connection

*1.* A first node is sampled [3] from the set of all **receiving nodes**.
*2.* A second node is sampled from the set of all **hidden** and **output nodes**.
*3.* A new weighted [4] connection is formed from the first to the second sampled node.

We give an example of this mutation in Figure 14.

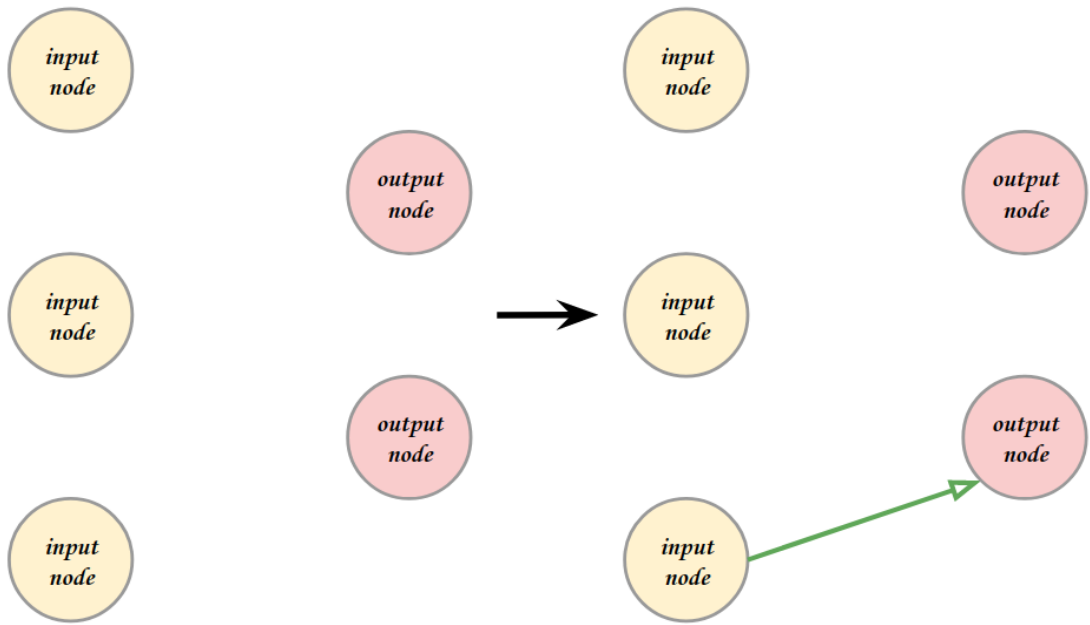

*Figure 14.* **DRN Mutation #1 : Grow Connection.** In order to showcase the four DRN mutations effectively, we are going to elaborate, over the next four figures, on a hypothetical scenario wherein we would like a neural network to process three scalar inputs and emit two scalar outputs. In this scenario, a DRN is to be initialized with three input nodes and two output nodes, all devoid of connections. Calling the "Grow Connection" mutation on such a DRN entails sampling a first node from the three input nodes and a second node from the two output nodes. Imagining that these turn out to be the third input node and the second output node, a weighted connection is therefore created in between these two nodes.

### 3.2.1.2. Mutation #2 : Prune Connection

*1.* A first node is sampled from the list of **emitting nodes**.
*2.* A second node is sampled from the first node's set of **out-nodes**.
*3.* The existing connection between the first and second node is subsequently deleted.

We give an example of this mutation in Figure 15.

---

[3] Sampling operations in these dynamic networks are always uniform.
[4] Where $\mathcal{N}(\mu, \sigma^2)$ is the Normal distribution with mean $\mu$ and variance $\sigma^2$, all DRN and DCN weights and biases are initialized with a value sampled from $\mathcal{N}(0, 1)$ and perturbed every iteration with a value sampled from $\mathcal{N}(0, 0.01)$.

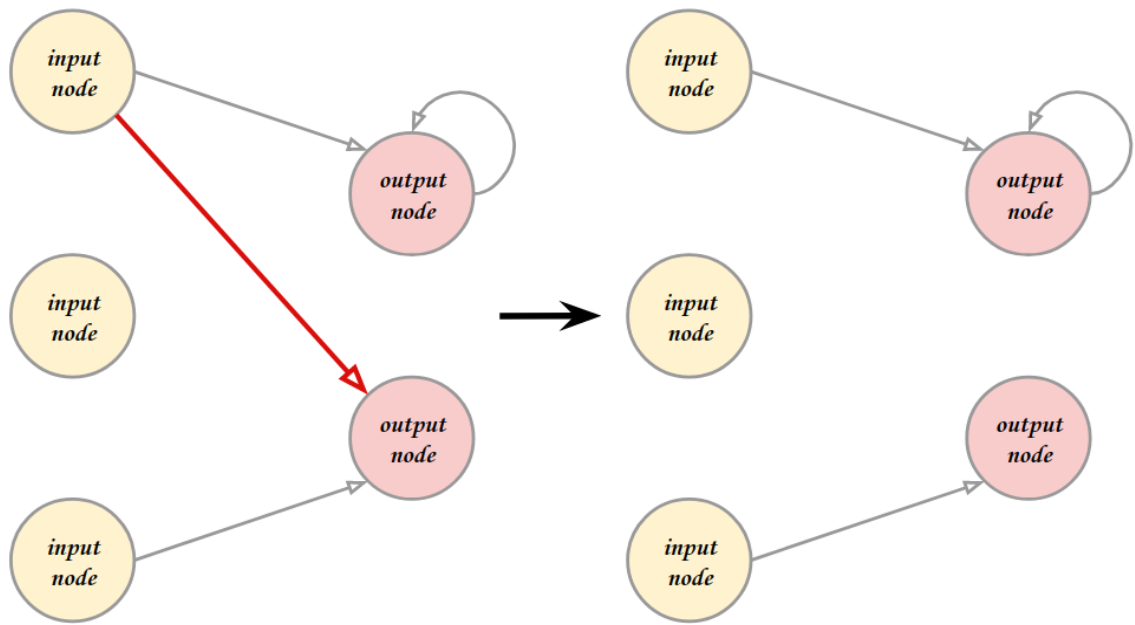

*Figure 15.* **DRN Mutation #2 : Prune Connection.** We pursue the scenario introduced in Figure 14 and pretend that three more "Grow Connection" mutations have been applied since then. Calling the "Prune Connection" mutation on this DRN entails sampling one node from a list composed of the first input node (twice), the third input node and the first output node. We imagine it turns out to be the first input node. A second node is now sampled from this node's set of out-nodes, which we in turn imagine to be the second output node. As a result, the connection between these two sampled nodes is deleted.

### 3.2.1.3. Mutation #3 : Grow Node

*1.* Three nodes are sampled:
- a first one, from the set of all **receiving nodes**.
- a second one, from the set of all **receiving nodes** minus the first sampled node.
- a third one, from the set of all **hidden** and **output nodes**.
*2.* A new hidden node is initialized in between.
*3.* Three connections are grown:
- from the first sampled node to the new hidden node.
- from the second sampled node to the new hidden node
- from the new hidden node to the third sampled node.

We give an example of this mutation in Figure 16.

### 3.2.1.4. Mutation #4 : Prune Node

*1.* A hidden node is sampled from the set of all **hidden** nodes.
*2.* The hidden node is deleted, alongside every of its connections.
Any other hidden node, which as a result of a DRN pruning mutation, is either not receiving or not emitting information is also deleted.

We give an example of this mutation in Figure 17.

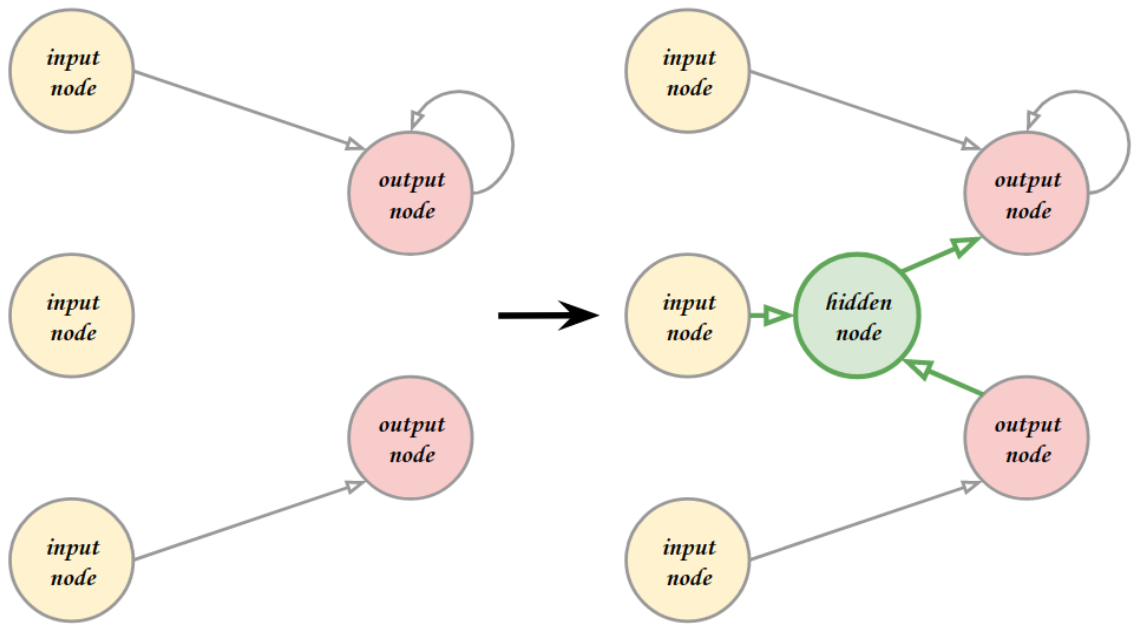

*Figure 16.* **DRN Mutation #3 : Grow Node.** We continue from where we left off in Figure 15. Calling the "Grow Node" mutation in this scenario entails sampling 1) a first node from all visible nodes 2) a second node from all the visible nodes excluding that first sampled node 3) a third node amongst the two output nodes. We imagine these to be the second input node, the second output node and the first output node respectively. A hidden node is created and connections are grown from 1) the second input node to the hidden node 2) the second output node to the hidden node 3) the hidden node to the first output node.

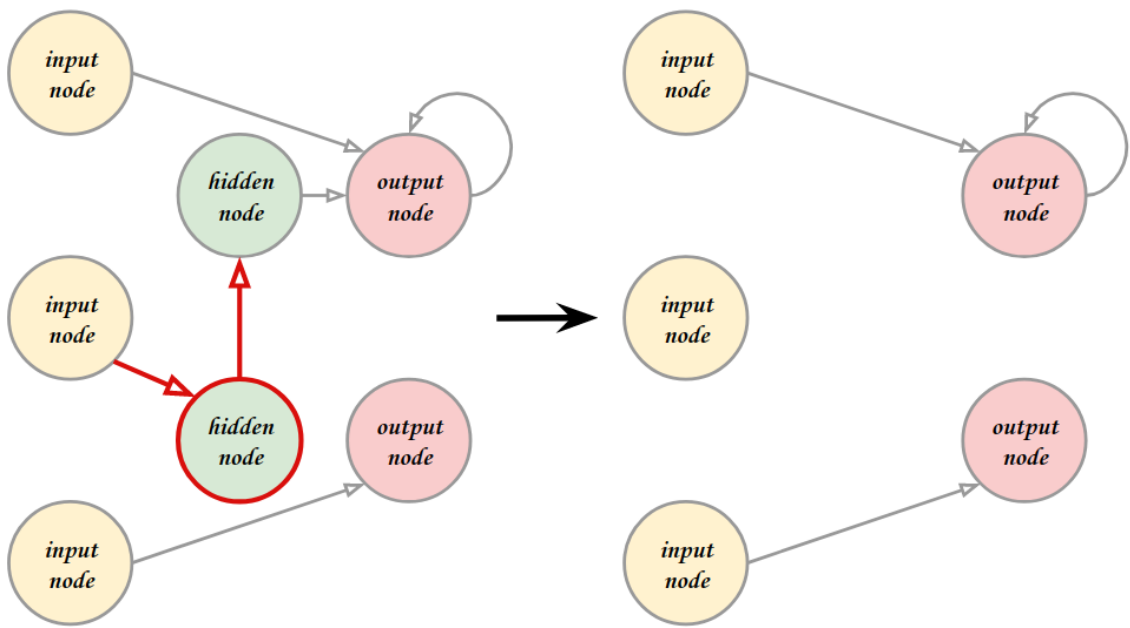

*Figure 17.* **DRN Mutation #4 : Prune Node.** We now imagine that many of the mutations described thus far have occurred since Figure 16. Calling the "Prune Node" mutation on this updated DRN entails sampling one node from the two hidden nodes. Imagining that it turns out to be the second hidden node, it is therefore deleted alongside its two connections. As a result of this deletion, the first hidden node, now devoid of any input information, is also deleted.

### 3.2.2. Dynamic Convolutional Network

We now describe a Dynamic Convolution Network (DCN). Connected through a tree-like structure, nodes in this network perform both convolution and pooling operations (Figure 18). Convolution nodes make use of standard unit-striding 3x3 kernels with variable amounts of input/output channels while pooling nodes use standard variable-sized square kernels with equivalent striding steps. Nodes outputting multi-valued feature maps are labeled hidden nodes while nodes outputting a singular value are given the title of output nodes. No padding operation is used except in the case of convolution nodes whose input feature maps are smaller than their 3x3 kernel size. Like the DRN, the DCN evolves in complexity according to four structural mutations:

1) Grow Branch (Section 3.2.2.2.)
2) Prune Branch (Section 3.2.2.3.)
3) Expand Node (Section 3.2.2.4.)
4) Contract Node (Section 3.2.2.5.)

We define the following terms:
**Branching nodes**: List of nodes from which "branches" have been grown from. Nodes appear in this list once per outcoming branch.
**Expanded nodes**: List of convolution nodes outputting more than one feature map. Nodes appear in this list once per additional output feature map.
**Pooling factor**: Factor by which a node reduces the dimensionality of its input through a pooling operation.

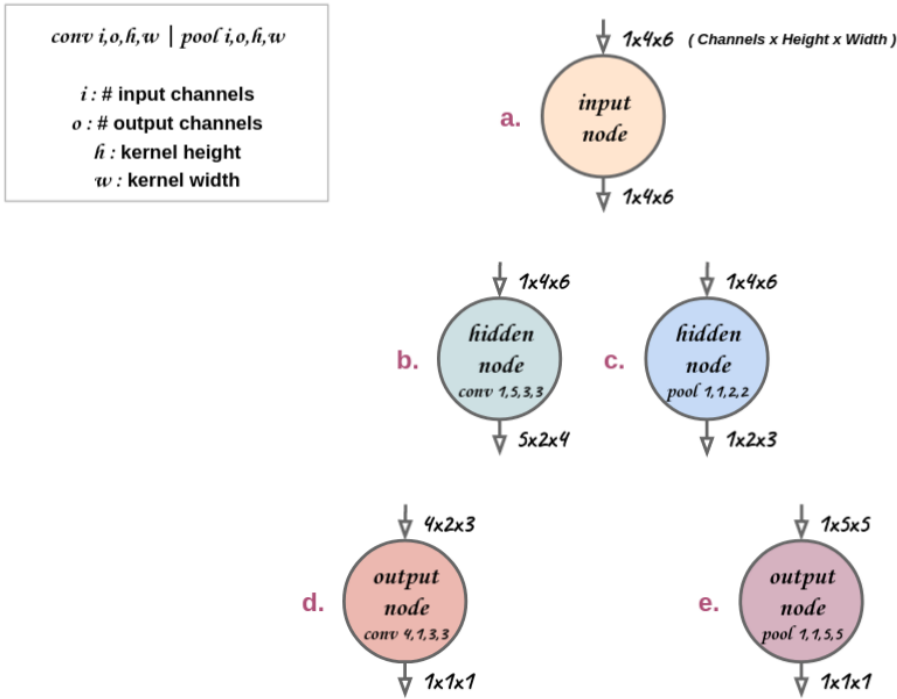

*Figure 18.* **Dynamic Convolutional Network Nodes.** Examples of the five types of nodes that ought to be encountered in a DCN. The root of a DCN tree is an input node (a). Internal/hidden nodes perform either convolution (b) or (average) pooling operations (c). Leaf/output nodes also perform either convolution (d) or (average) pooling operations (e), but unlike hidden nodes, always output one single value. In these networks, information flows from top to bottom without recurrence.

### 3.2.2.1. Grow Node Operation

In a DCN, all nodes (except the root node) have one parent node and potentially many children nodes. New DCN nodes are always created through the "Grow Node" operation. This operation creates a new node and appends it to an existing node's set of out-nodes. While nodes that perform pooling operations always grow child nodes that perform convolution operations, the following child-growing process takes place for input and convolution nodes. In order to determine a new node's function (convolving or pooling), a pooling factor, based on the parent node's output dimensionality is determined. Originally, the first new child node will be given the largest possible pool factor which correctly divides the parent node's output feature map. Then, once a child pools by this factor, the next largest one will be given to its sibling and so on, until no pooling factor is available in which case the child is determined to be a convolution node.

We give an example of this procedure in Figure 19.

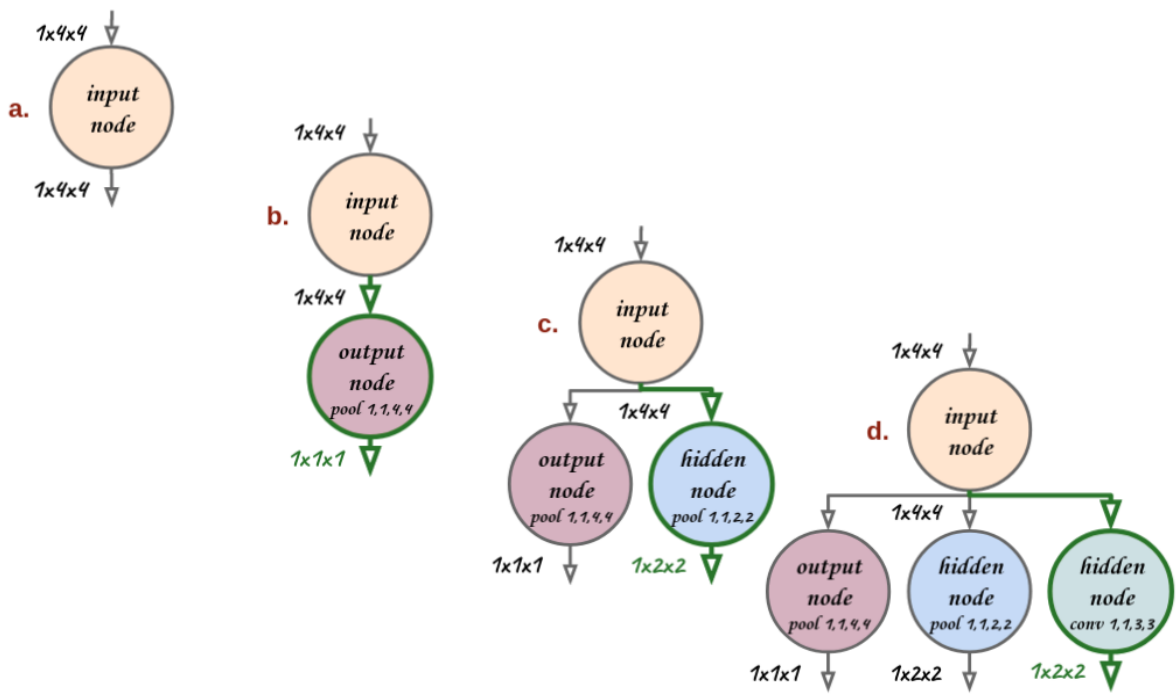

*Figure 19.* **DCN Operation : Grow Node.** To illustrate the process of growing new nodes, we set up a DCN input node that processes input matrices of size 1x4x4 (a). At this point in time, calling the "Grow Node" operation on this input node results in creating a new node with the pooling factor that best divides the input's node output: 4, which leads to the creation of an output pooling node with such factor (b). Calling the "Grow Node" operation once again on the input node will result in creating a new node with the second biggest possible pooling factor: 2, which leads to the creation of a hidden pooling node (d). Finally, since we've already used up every pooling factor, calling the "Grow Node" operation one last time results in the creation of a hidden convolution node.

### 3.2.2.2. Mutation #1 : Grow Branch

*1.* A base node is sampled from the set composed of the **input node** and all **hidden nodes**.
*2.* The "Grow node" operation is called on this base node.

*3.* Until the "Grow node" operation yields an **output node**, it is repeatedly called on the most recently created node.

We give an example of this mutation in Figure 20.

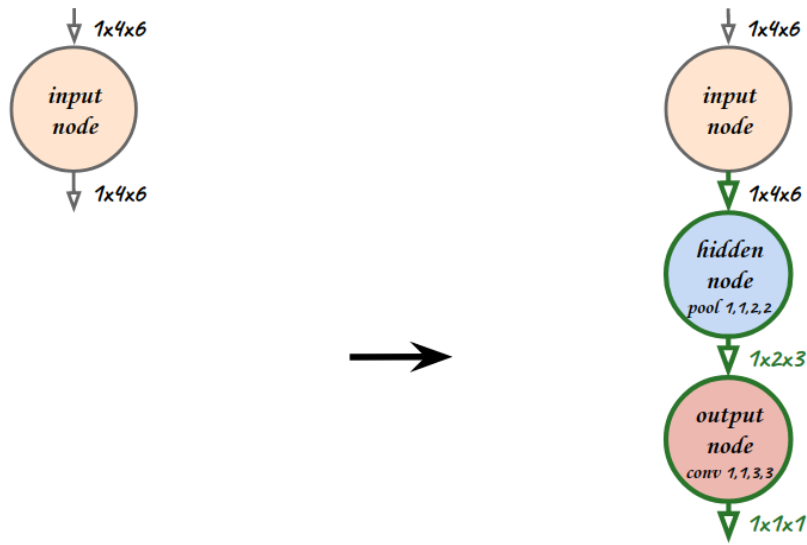

*Figure 20.* **DCN Mutation #1 : Grow Branch.** In order to showcase the four DCN mutations effectively like we did for the DRN, we are going to iterate over the next four figures on a scenario wherein we require a network to process some image encoded in a 1x4x6 matrix. In order to receive this information, the DCN is first of all initialized with a single corresponding input/root node. At this point, since the tree is solely composed of this one node, calling the "Grow Branch" mutation automatically selects it. As it is devoid of out-nodes, calling this "Grow Branch" mutation on the input node results in the creation of a new pooling node with the biggest possible pooling factor: 2. With this pooling factor, this newly created node is to output 1x2x3 matrices, marking it as a hidden node. A new "Grow Node" operation is therefore called once again, this time on the newly created hidden pooling node. Since this node is a pooling node, its child is made by default to be a convolution node. With this convolution operation, this latest node is to output 1x1x1 matrices, marking it as an output node, signaling the end of the "Grow Branch" mutation.

### 3.2.2.3. Mutation #2 : Prune Branch

*1.* A first node is sampled from the list of all **branching nodes**.
*2.* A second node is sampled from the first node's set of ***out-nodes***.
*3.* This second node, in addition to the entirety of its sub-tree, is subsequently pruned.

We give an example of this mutation in Figure 21.

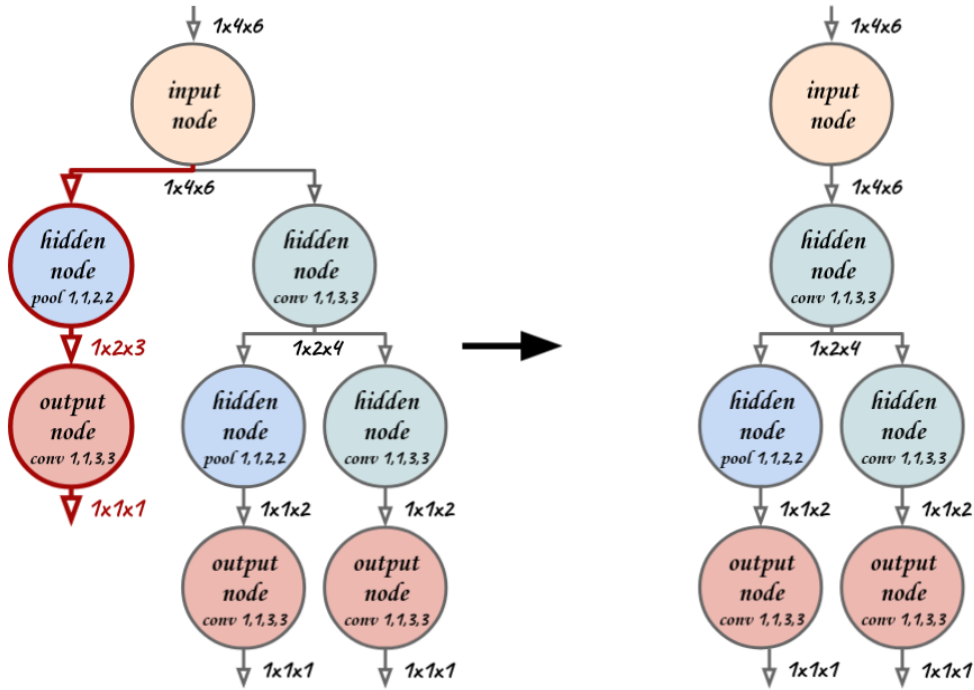

*Figure 21.* **DCN Mutation #2 : Prune Branch.** We continue from the scenario introduced in Figure 20 and assume two more "Grow Branch" mutations have been performed: one on the input node and another on the resulting second layer hidden convolution node. Calling the "Prune Branch" mutation on this DCN entails sampling a first node from the list composed of: the input node (twice) and the second layer hidden convolution node. We imagine it turns out to be the first input node and so a second node is now sampled from this input node's set of out-nodes, which we imagine to be the second layer hidden pooling node. As a result, this hidden pooling node and its subtree (composed of one output convolution node) are deleted.

### 3.2.2.4.  Mutation #3 : Expand Node

*1.* A node is sampled from the set of **convolution nodes** that are neither **output nodes** nor have an **out-node** that is an **output pooling node**.
*2.* A new series of weights is randomly initialized for this node to output an additional feature map from its input.
*3.* This node's convolution children (and grandchildren for each pooling child) make the necessary adjustments (create a new set of weights) in order to input this additional feature map.

We give an example of this mutation in Figure 22.

### 3.2.2.5.  Mutation #4 : Contract Node

*1.* A node is sampled from the list of **expanded** nodes.
*2.* One of this node's output weight channels is sampled and deleted. This node's children's (and grandchildren's for each pooling child) corresponding input weight channels are subsequently deleted.

We give an example of this mutation in Figure 23.

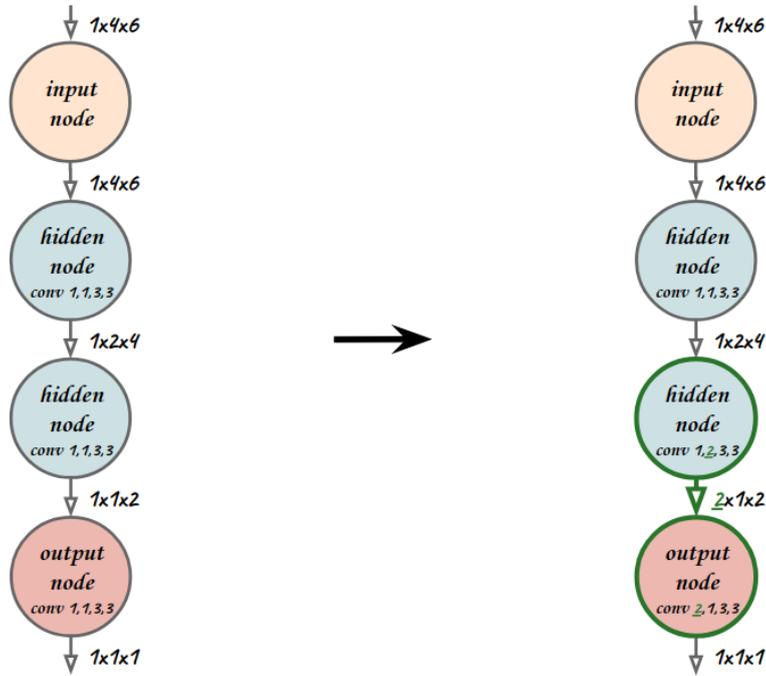

*Figure 22.* **DCN Mutation #3 : Expand Node.** We continue from where we left off in Figure 21 and assume one "Prune Branch" mutation has been called on the second layer hidden node, thereafter deleting its left subtree. Calling the "Expand Node" mutation on this DCN first entails sampling from a set composed of both hidden nodes. We imagine it to be the third layer hidden node. A new series of weights is initialized inside this node to output a second feature map. Its out-node, the output convolution node, initializes new weights to be able to utilize this new feature map.

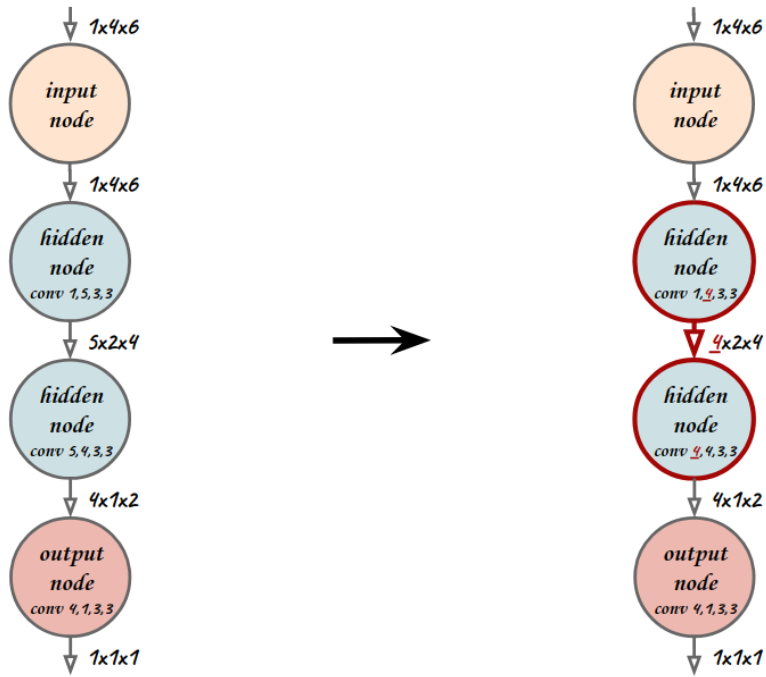

*Figure 23.* **DCN Mutation #4 : Contract Node.** For this last mutation, we assume that, since Figure 22, six new "Expand Node" mutations have been called: four on the second layer hidden node and two on the third layer hidden node. Calling the "Contract node" mutation on this DCN entails sampling one node from the list composed of the second layer hidden node (four times) and the third layer hidden node (three times). We imagine it turns out to be the second layer hidden node. As a result, one of its output weight channels is randomly deleted. Its out-node, the third-layer convolution node, deletes its corresponding weights.

### 3.2.3. Connecting the Dynamic Networks

Due to the single-valued nature of its outputs, the Dynamic Convolutional Network (DCN) can be made to connect to a Dynamic Recurrent Network (DRN). In order to handle the dynamic number of leaf nodes in a DCN resulting from its "Grow Branch" and "Prune Branch" mutations, the DRN needs to handle a variable amount of input nodes. To do so, the following operations take place :

- Following the call of a DCN "Grow Branch" mutation, the DRN grows a corresponding input node, subsequently connected to a random hidden or output node (Figure 24).

- Following the call of a DCN "Prune branch" mutation, the DRN prunes the corresponding (potentially many) input nodes, along with their connections and hidden nodes now devoid of input information (Figure 25).

Connecting these two dynamic networks results in eight structural mutations. Regardless of the number of connected networks, every iteration, one mutation is sampled from the set of all potential mutations and subsequently applied to the corresponding sub-network.

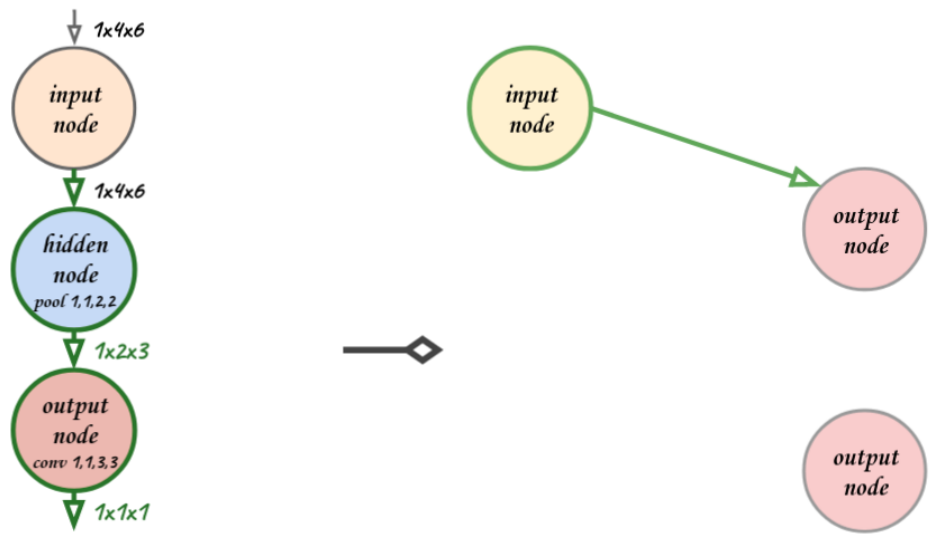

*Figure 24.* **Connecting the Dynamic Networks : Grow Branch Mutation.** Over the next two figures, let us introduce a revised hypothetical scenario in which we would like a neural network to input a 1x4x6 image and output two scalar values. To do so, we initialize a DCN to receive this input matrix and a DRN to output these two scalars. Similarly to Figure 20, the DCN is initialized with an input node whereas unlike Figure 14, the DRN only initializes the two output nodes. Now, once again, calling the "Grow Branch" mutation on the DCN results in a series of nodes being created until the creation of an output/leaf node. As a result of this new output DCN node, the DRN creates an input node of its own to receive the value outputted by that leaf node. Following the creation of this input node, a random node is sampled from the set of all hidden and output nodes, which in this case we assume to be the first output node, for the input node to finally connect to.

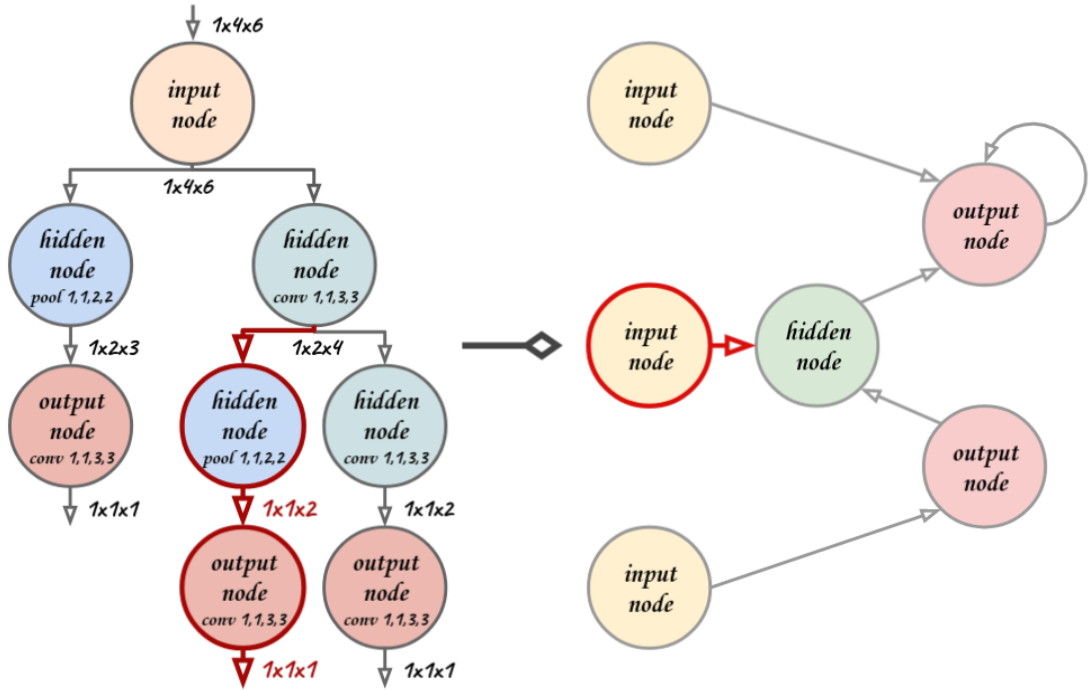

*Figure 25.* **Connecting the Dynamic Networks : Prune Branch Mutation.** Let us now assume that many mutations have occurred since Figure 24 inside both sub-networks, resulting in three DCN-DRN connections. In this configuration, in the event that a "Prune Branch" mutation occurred onto the 3rd layer hidden pooling node of the DCN, not only would the DCN branch be deleted, but the corresponding second DRN input node would also be pruned with its single connection.

### 3.3. Scalable Communication Protocol for Evolutionary Algorithms

As mentioned earlier, evolutionary algorithms generally require setting up a population of agents through an iterative three-stage process; these stages being variation, evaluation and selection. Naturally, increasing the number of agents in the population increases the computational cost of each of these stages. Most often however, the operations performed within each stage can be parallelized. When agents do not interact, variation and evaluation can even be performed completely independently for each agent. In that case, the workload ought to be dispatched to separate processes to increase efficiency.

Such et al., 2017 recently introduced a technique wherein variations ought to be encoded as "seeds". In that framework, each of these seeds encodes for a perturbation applied to all of the weights of a standard deep neural network, and agents as they evolve, are continuously appended new seeds representing new perturbations, stacked on top of each other.

This idea is most easily implemented by splitting processes into one main process and many side processes. Every iteration, the main process generates new seeds and scatters them to side processes who then go on to build and evaluate the agents they are assigned. Once done, the main process gathers back the fitnesses computed for all agents in order to select seeds for the next generation to build on (Figure 26). However, with this framework, processes are required to rebuild agents from scratch every generation, resulting at best in a continuous linear cost increase of the variation stage. When variation operations are cheap, this increase in cost is insignificant. However this approach quickly loses feasibility as the computational cost of variation operations increases.

We therefore propose an alternative implementation which capitalizes on a 50% duplicative truncation selection. We set up a communication scheme wherein, every generation, each process in possession of a selected agent communicates it over to a process in possession of a non-selected agent (Figure 27). With this approach, at iteration *n*, processes replace *n* variation operations with *1* communication operation and *1* variation operation. Due to high communication bandwidth between processes in modern computer clusters, this communication operation tends to be virtually free, giving way to a much more scalable evolution process.

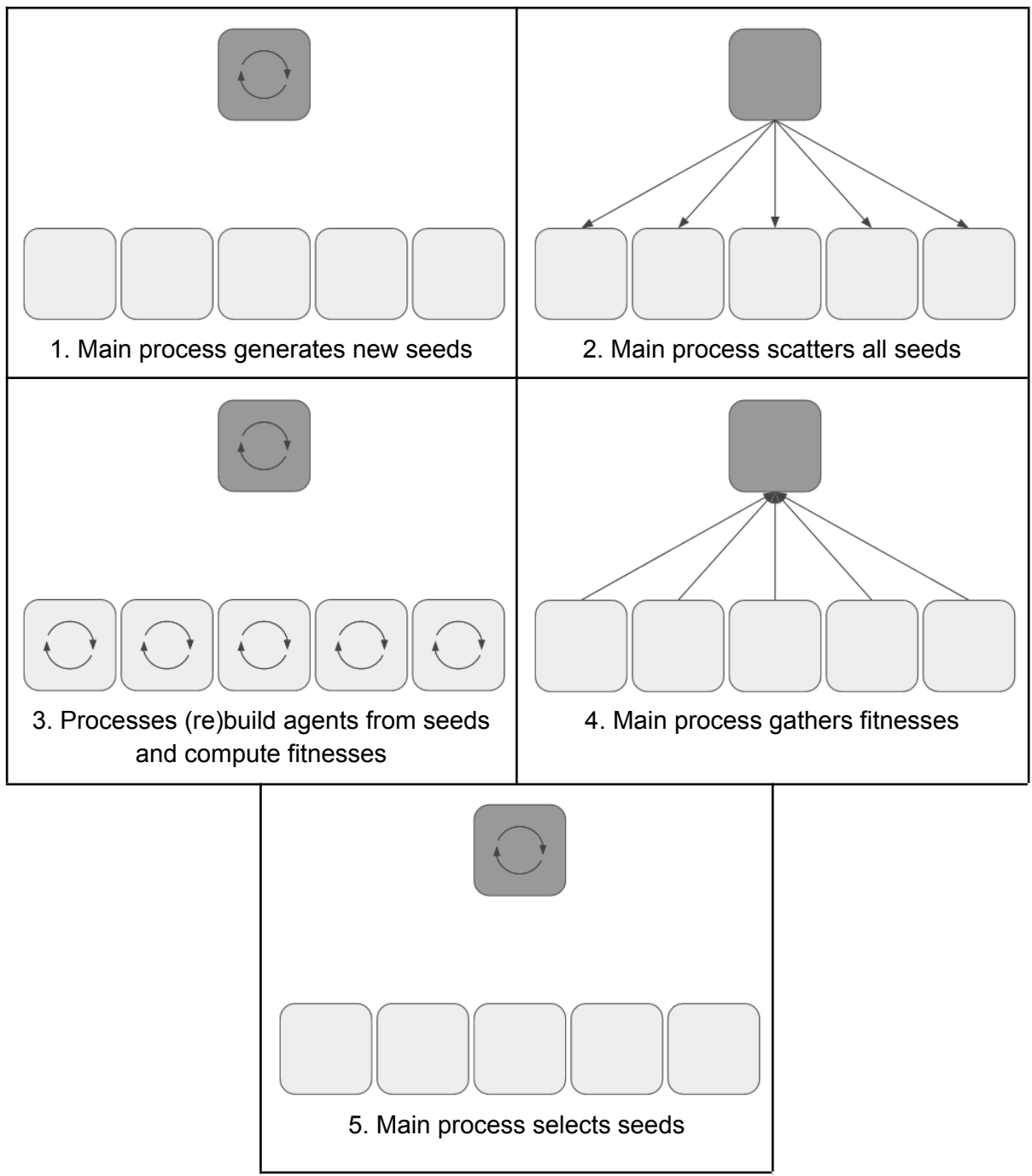

1. Main process generates new seeds
2. Main process scatters all seeds
3. Processes (re)build agents from seeds and compute fitnesses
4. Main process gathers fitnesses
5. Main process selects seeds

*Figure 26.* **Evolutionary Inter-Process Communication : Scatter/Gather.**

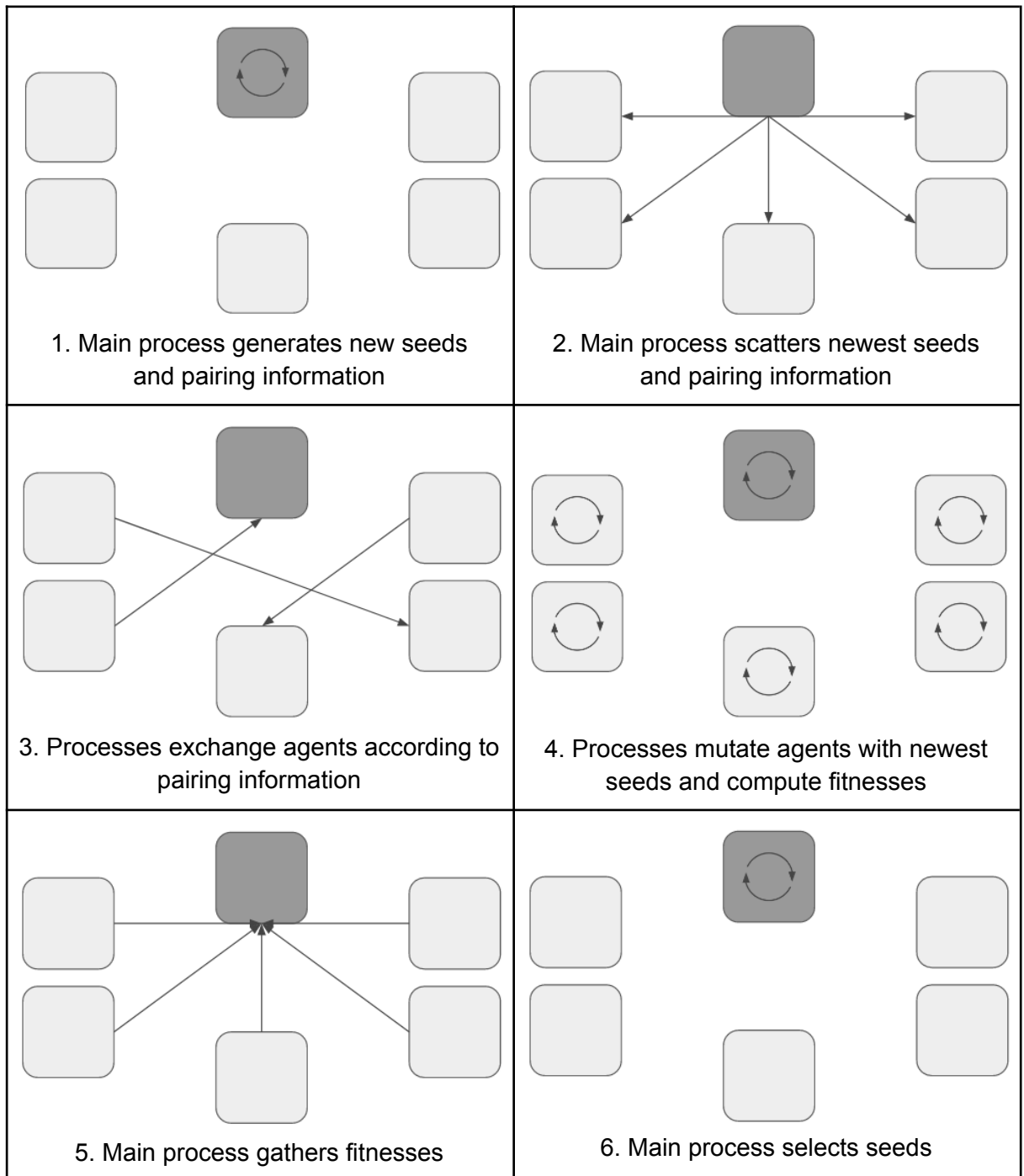

*Figure 27.* **Evolutionary Inter-Process Communication : Scatter/Gather + Peer-to-Peer.**

## 4. Experiments

### 4.1. Score Optimization and Deep Reinforcement Learning Agent Imitation

We evolve, on a variety of tasks, both score-maximizing agents, selected based on the number of points earned in a single episode, and agents imitating pre-trained Deep Reinforcement Learning agents (Raffin et al., 2019), selected in the context of the presented adversarial generation framework. We evaluate the various action-taking agents by looking at the mean score obtained across populations of various sizes over five newly seeded runs.

### 4.1.1. CartPole and LunarLander

We first evaluate our methods on CartPole and LunarLander, tasks for which state observations fed to neural networks are encoded into sets of scalar values. We attempt to imitate, for the adversarial generation portion of the experiment, PPO (Schulman et al., 2017) and TQC (Kuznetsov et al., 2020) agents respectively. We experiment with two types of networks: Dynamic Recurrent Networks (DRN) and the three-layer feedforward neural networks used by the Deep RL agents (except in discriminators for which we turn the last hidden fully-connected layer into a recurrent one). We find that both score maximizing dynamic and static networks are able to solve the tasks. Furthermore, agents optimized for imitation turn out quite capable and getting increasingly better as population size increases.

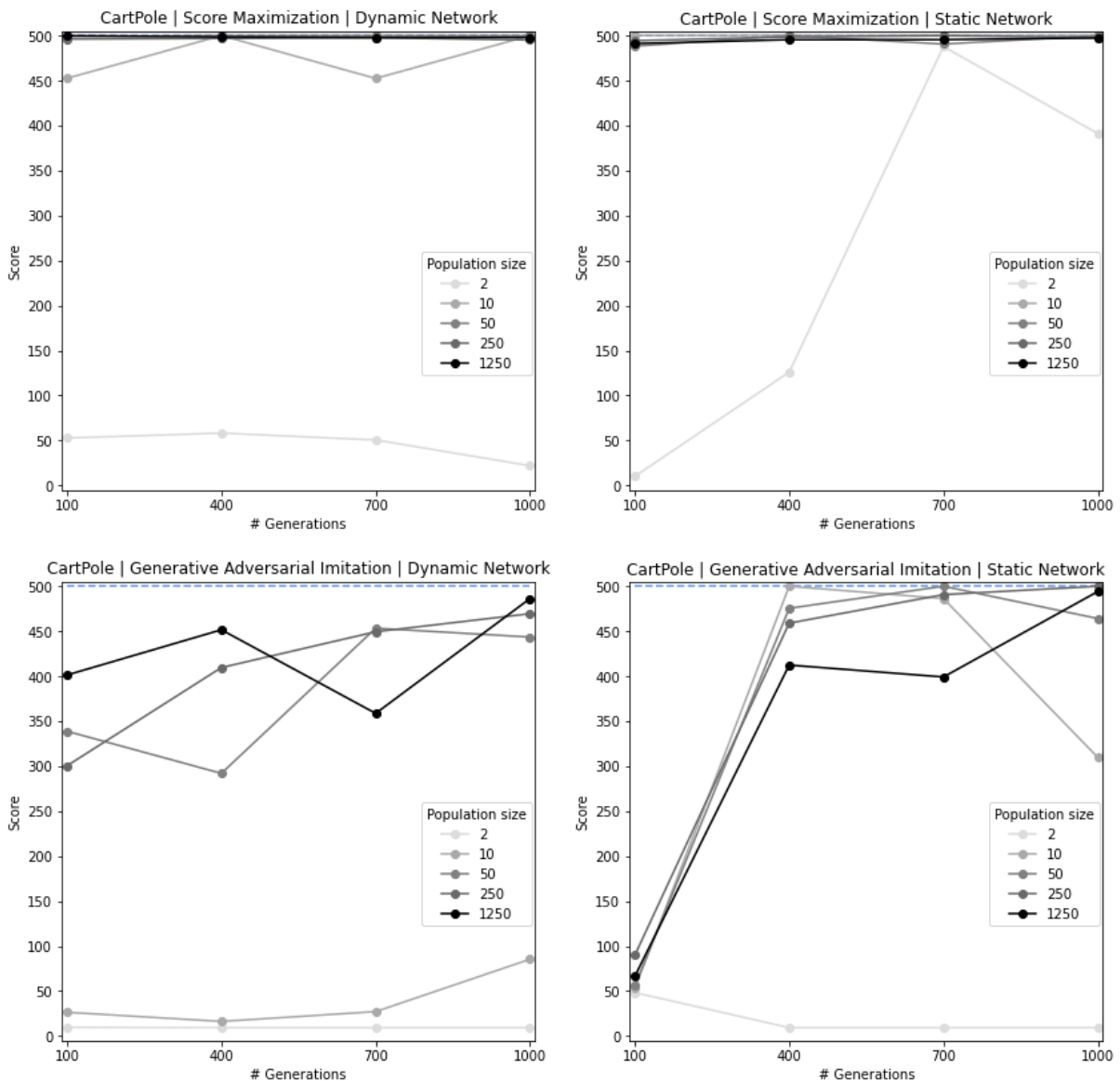

*Figure 28.* **Score Maximization and Generative Adversarial Imitation on CartPole.** The dotted blue line is the mean score obtained by PPO, which is also the highest achievable score on this task. Starting from a population size of 10, both dynamic and static score-optimizing agents also consistently achieve the highest score. Imitating agents for both types of networks, starting from a population size of 50, also make their way up to the highest score.

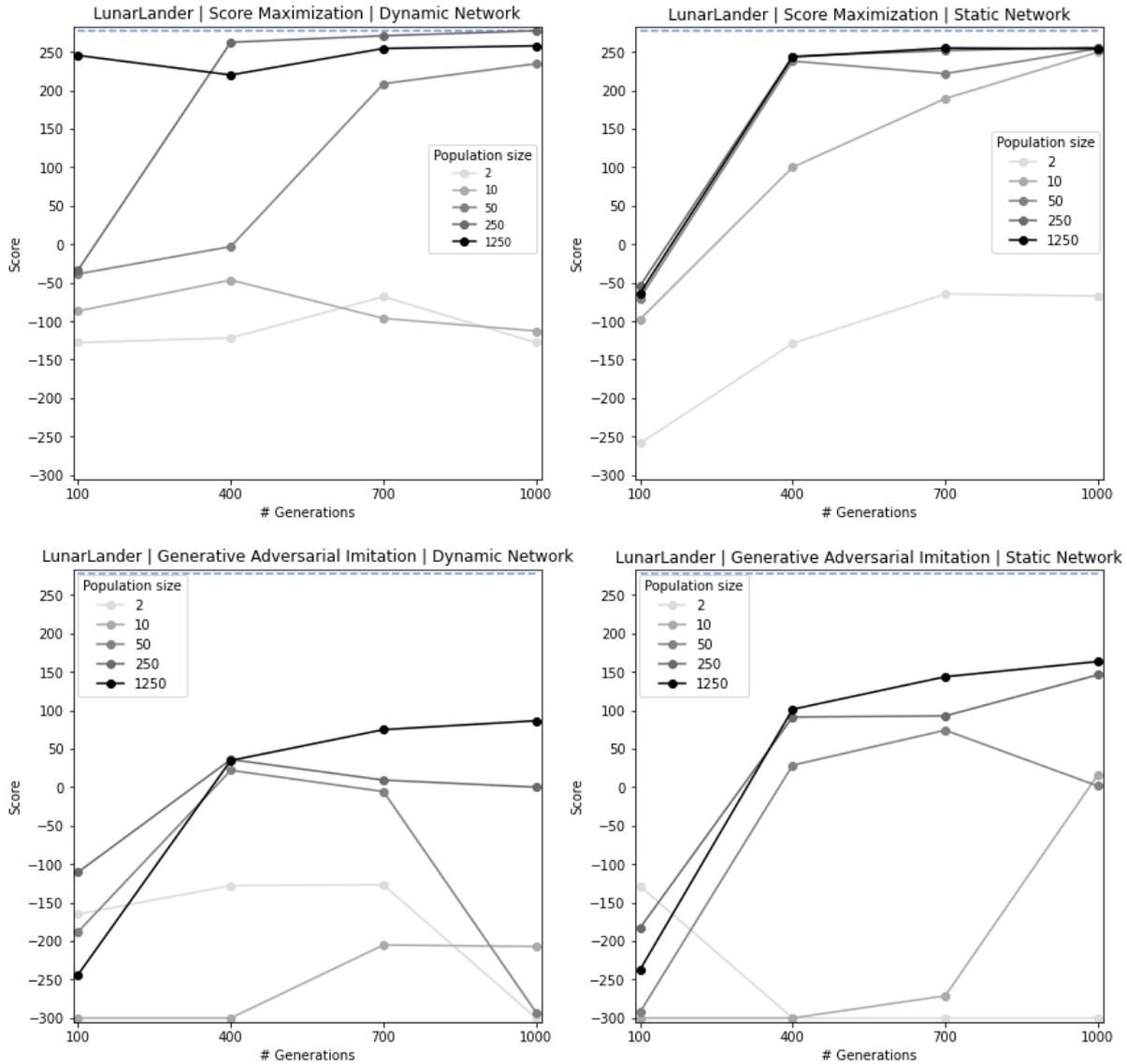

*Figure 29.* **Score Maximization and Generative Adversarial Imitation on LunarLander.** The dotted blue line is the mean score obtained by TQC. Both score-optimizing dynamic and static networks achieve similar performance as the state-of-the-art baseline starting from population sizes of 50 and 10 respectively. Action-taking agents optimized through adversarial generation, for both dynamic and static networks, clearly increase their performance as iterations and population size increases.

### 4.1.2. Pong

We now experiment with Pong, a task that Deep RL agents have historically performed remarkably well on (Mnih et al., 2013; Schulman et al., 2017) but that Deep GA agents have not (Such et al., 2017). We move on to using dynamic networks combining a Dynamic Convolutional Network (DCN) and a Dynamic Recurrent Network (DRN), which we match against 100k parameters deep convolutional recurrent networks. As opposed to the two previous tasks wherein whole episodes last no more than seconds, Pong games can last a few minutes. As a proof of concept we therefore propose to only let agents play the first 400 frames (100 actions, ~7 seconds). Finally, we attempt, for the adversarial generation section of the experiment, to imitate a PPO (Schulman et al., 2017) agent. We find that dynamic networks perform much better than static networks on the score-optimization task, and while all agents optimized to imitate do not perform as well as score-maximizing agents after 1000 generations, dynamic networks show clear signs of improvement as population size increases.

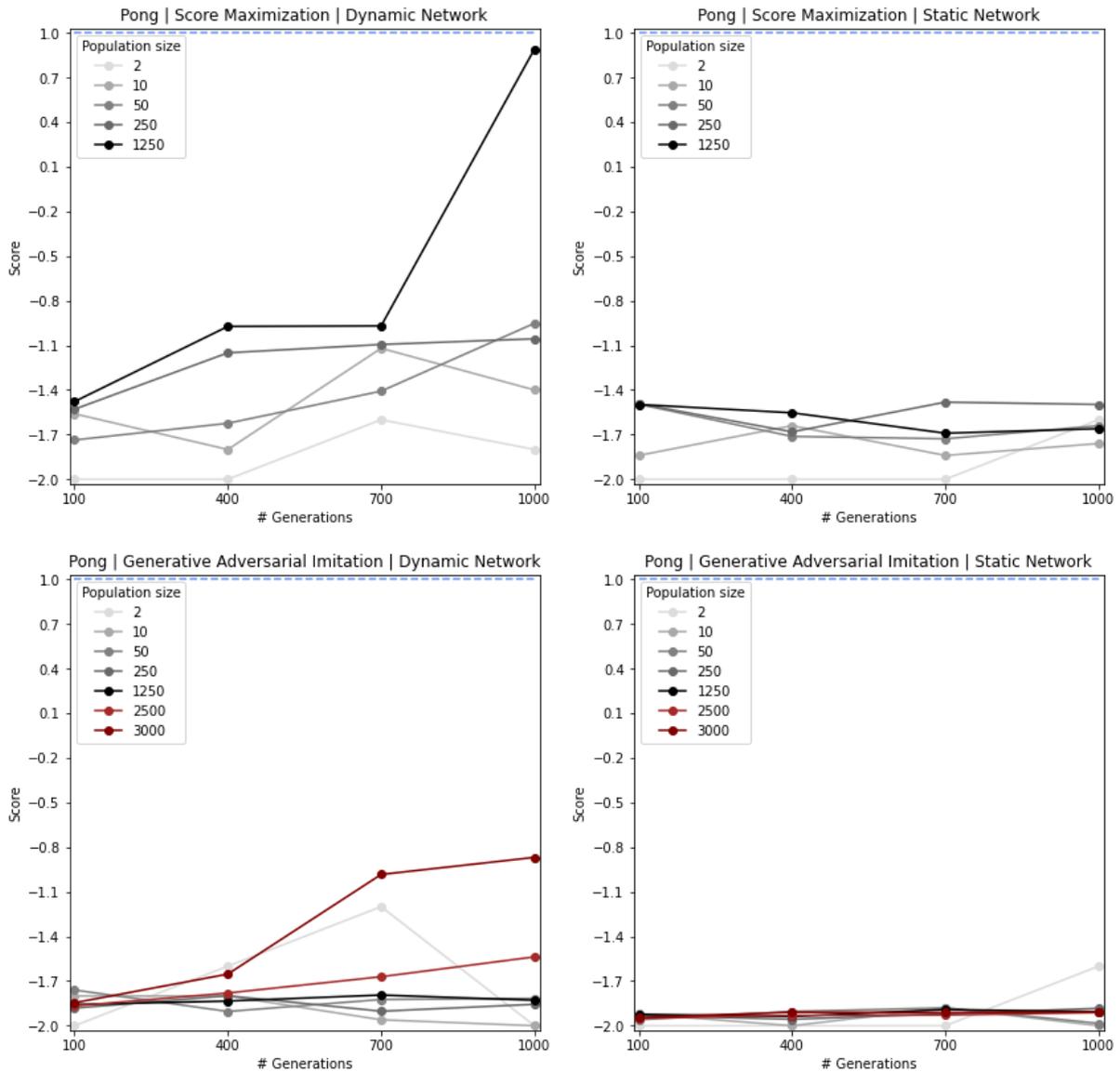

*Figure 30.* **Score Maximization and Generative Adversarial Imitation on Pong.** The dotted blue line is the mean score obtained by PPO. Dynamic networks fare much better than static networks after 1000 generations on pure score maximization, reaching an average of 0.89 points across the population of size 1250. Action-taking agents optimized through adversarial generation perform much worse after the first 1000 generations than their score-optimizing counterparts, however, dynamic agents appear to be trending upwards as population sizes get bigger.

### 4.2. Human Behaviour Imitation on Shinobi III : Return of the Ninja Master

We now move on to attempting to create agents with human-like behaviour in a video game context. We make use of behavioural data collected through gym-retro (Nichol et al., 2018), in the context of the Courtois Neuromod project (Boyle et al., 2019), on the video game *Shinobi III: Return of the Ninja Master* [5]. This game is played with a controller (Figure 31) through which the player has access to a variety of actions and combinations of actions in order to move and interact with non-player characters (Table 1). Random and score-optimizing agents produce behaviour displayed in Table 2. Our subjects however, happen to not be trying to optimize score, leading to behaviour dissimilar to the two prior examples (Table 3).

---

[5] The protocol for the experiments was approved by the local ethics institutional review board, the "Comité d'éthique de la recherche vieillissement-neuroimagerie" and all participants provided informed consent to participate in the study.

We once again make use of the evolutionary adversarial generation framework and equip generators and discriminators alike with a DCN feeding into a DRN while also running experiments for similar hundred thousand parameter deep neural networks.

In order to quantify the quality of these agents, we first develop a qualitative behaviour assessment table for each of our subjects (Table 4). We then come up with qualitative assessments for imitation quality displayed by the agents (Table 5). Using these, we show our results in Table 6. We find that given sufficient computational resources, agents embedded with static and dynamic networks all display behaviours showing similarity with the corresponding human subject. However, agents operating with networks of dynamic complexity appear to better leverage the increase in computation, showcasing for sub-02 almost indistinguishable behaviour. In this particular case, both sub-networks were composed of ten layers each, made use of around one hundred nodes and exchanged information through forty DCN output nodes.

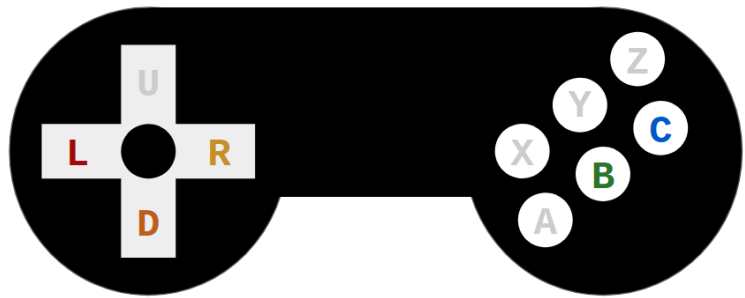

*Figure 31.* **SEGA Genesis controller diagram.**

| R | walk right | video |
|---|---|---|
| L | walk left | video |
| D | crouch | video |
| C | jump | video |
| B | *(no enemy nearby)* throw a projectile | video |
| B | *(enemy nearby)* swing a blade | video |
| R, pause, R | run | video |
| C, pause, C | double jump | [video](#) |
| C, pause, B | jump and throw a projectile | [video](#) |
| R, pause, R+B | run and swing a blade | [video](#) |
| C, pause, D+B | jump down kick | [video](#) |
| C, pause, C+B | double jump and throw projectiles | [video](#) |

Table 1. **Video samples of possible actions and combinations of actions in Shinobi III.**

| random agent | score-optimizing agent |
|---|---|
| [video](#) | [video](#) |

Table 2. **Video samples of random and score-optimizing agents in Shinobi III.**

| sub-01 | | | | | sub-02 | | | | |
|---|---|---|---|---|---|---|---|---|---|
| [video](#) | [video](#) | [video](#) | [video](#) | [video](#) | [video](#) | [video](#) | [video](#) | [video](#) | [video](#) |

Table 3. **Video samples of subjects' behaviour in Shinobi III.**

| | sub01 | sub02 |
|---|---|---|
| **elements of behaviour** | - moves right<br>- doesn't throw projectiles<br>- jumps<br>- down kicks after jumping<br>- runs<br>- swings a blade after running | - moves right<br>- doesn't throw projectiles<br>- doesn't jump<br>- runs<br>- swings a blade after running |
| **behaviour** | 1) moves right by running or walking right, 1x jump down kick, walking right<br>2) takes down first character with 1x swing or 1x jump down kick<br>3) moves right by running or walking right, 1x jump down kick, walking right<br>4) ignores the second character in the tree<br>5) takes down third character with 1x swing or 2x jump down kick | 1) moves right by running<br><br>2) takes down first character with 1x swing<br><br>3) moves right by running<br><br>4) ignores the second character in the tree<br>5) takes down third character with 1x swing |

Table 4. **Qualitative assessment table of subjects behaviour in Shinobi III.**

| | | Agent behaviour indistinguishable from subject behaviour |
|---|---|---|
| | | Agent behaviour close to indistinguishable from subject behaviour |
| | | Agent behaviour very similar to subject behaviour but definitely distinguishable |
| | | Agent behaviour mostly similar to subject behaviour |
| | | Agent behaviour becoming similar to subject behaviour |
| | | Agent shows clear elements of subject behaviour |
| | | Agent appears to be showing multiple elements of subject behaviour |
| | | Agent appears to be showing at least two elements of subject behaviour |
| | | Agent behaving almost completely differently from subject |
| | | Agent behaving completely differently from subject |

*Table 5.* **Qualitative assessment table of agent behaviour in Shinobi III.** In order to evaluate our agents on their ability to imitate our subjects in the first few seconds of gameplay, we develop a qualitative metric of performance based on the observations made in Table 4.

| | time ↓ | sub-01 | | | | sub-02 | | | |
|---|---|---|---|---|---|---|---|---|---|
| pop size → | | 10 | 40 | 160 | 640 | 10 | 40 | 160 | 640 |
| static | 12h | video | video | video | video | video | video | video | video |
| static | 24h | video | video | video | video | video | video | video | video |
| static | 36h | video | video | video | video | video | video | video | video |
| static | 48h | video | video | video | video | video | video | video | video |
| dynamic | 12h | video | video | video | video | video | video | video | video |
| dynamic | 24h | video | video | video | video | video | video | video | video |
| dynamic | 36h | video | video | video | video | video | video | video | video |
| dynamic | 48h | video | video | video | video | video | video | video | video |

*Table 6.* **Table of Shinobi III imitation results.** We experiment with four population sizes (one agent per process) for 48 hours each. To evaluate the dynamic architectures we also run experiments with standard static deep neural networks. Each video sample presented here was collected by looking at the first three generators in the population and selecting the one we deemed closest in behaviour to the corresponding subject. We include color codes corresponding to our qualitative assessment table presented in Table 5.

## 5. Discussion

As a result of these experiments, we believe that, given sufficient computational resources, it is indeed possible to successfully evolve artificial agents to adopt behaviours similar to human subjects in video game environments. The adversarial generation framework appears well-suited to evolve artificial agents towards being able to encode the complex functions that are: playing similarly to a human player and differentiating human play to other types of behaviours. Just as important, these two complex functions appear to be encodable in the proposed neural networks of dynamic complexity. Finally, we would like to point out the value of the peer-to-peer communication protocol, which made it realistic to evolve these agents for tens of thousands of generations as opposed to just hundreds.

While this study only showcases the modeling of short behavioural sequences, these preliminary results offer confidence in these techniques' ability to model longer timeframes. Finally, this work further suggests that, powered by increasingly strong resources, evolutionary computation could be capable of tackling even more complicated problems in the future.

## 6. Conclusion

This paper introduced three separate evolutionary computation techniques: an evolutionary adversarial generation framework, a procedure to develop artificial neural networks of dynamic complexity and a scalable inter-process communication protocol. Combining these techniques yields, under sufficient computing resources, artificial video game behaviour quasi-identical from human subjects on short temporal windows. These results further demonstrate that evolutionary optimization can be leveraged to solve complex high-dimensional problems.